\documentclass[11pt,a4paper]{article}
\usepackage[hyperref]{emnlp2020}
\usepackage{times}
\usepackage{latexsym}
\usepackage{multirow}
\usepackage{amssymb}
\usepackage{graphicx}
\usepackage{amsmath}
\usepackage{subfigure}
\usepackage{makecell}
\usepackage{booktabs}
\usepackage{enumitem}

\usepackage{microtype}

\aclfinalcopy 
\def\aclpaperid{782} 

\title{Is {G}raph {S}tructure {N}ecessary for {M}ulti-hop {Q}uestion {A}nswering?}

\author{
	Nan Shao$^\dag$,
  	Yiming Cui$^\ddag$$^\dag$,
  	\textbf{Ting Liu$^\ddag$},
  	\textbf{Shijin Wang$^\dag$$^\S$},
  	\textbf{Guoping Hu$^\dag$}\\
  	{$^\dag$State Key Laboratory of Cognitive Intelligence, iFLYTEK Research, China} \\
  	{$^\ddag$Research Center for Social Computing and Information Retrieval (SCIR), } \\
  	{Harbin Institute of Technology, Harbin, China} \\
  	{$^\S$iFLYTEK AI Research (Hebei), Langfang, China} \\
  	$^\dag$$^\S$\tt\{nanshao,ymcui,sjwang3,gphu\}@iflytek.com \\
  	$^\ddag$\tt\{ymcui,tliu\}@ir.hit.edu.cn
  }

\date{}

\begin{document}
\maketitle

\begin{abstract}
Recently, attempting to model texts as graph structure and introducing graph neural networks to deal with it has become a trend in many NLP research areas.
In this paper, we investigate whether the graph structure is necessary for multi-hop question answering.
Our analysis is centered on HotpotQA. 
We construct a strong baseline model to establish that, with the proper use of pre-trained models, graph structure may not be necessary for multi-hop question answering.
We point out that both graph structure and adjacency matrix are task-related prior knowledge, and graph-attention can be considered as a special case of self-attention.
Experiments and visualized analysis demonstrate that graph-attention or the entire graph structure can be replaced by self-attention or Transformers.
\end{abstract}

\section{Introduction}
\label{sec:introduction}

Different from single-hop question answering, where the answer can be derived from a single sentence in a single paragraph, more and more studies focus on multi-hop question answering across multiple documents or paragraphs \citep{welbl-etal-2018-constructing,talmor-berant-2018-web,yang-etal-2018-hotpotqa}. 

To solve this problem, the majority of existing studies constructed a graph structure according to co-occurrence relations of entities that scattered across multiple sentences or paragraphs. \citet{dhingra-etal-2018-neural} and \citet{song2018exploring} designed a DAG-styled recurrent layer to model the relations between entities. \citet{de-cao-etal-2019-question} first used GCN \citep{kipf2017semi} to tackle entity graph. \citet{qiu-etal-2019-dynamically} proposed a dynamic entity graph for span-based multi-hop QA.
\citet{tu-etal-2019-multi} extended the entity graph to a heterogeneous graph
 by introducing document nodes and query nodes.

Previous works argue that a fancy graph structure is a vital part of their models and demonstrate that by ablation experiments. 
However, in experiments, we find when we use the pre-trained models in the fine-tuning approach, removing the entire graph structure may not hurt the final results.
Therefore, in this paper, we aimed to answer the following question: 
\textbf{How much does graph structure contribute to multi-hop question answering?}

To answer the question above, we choose the widely used multi-hop question answering benchmark, HotpotQA \citep{yang-etal-2018-hotpotqa}, as our testbed. 
We reimplement a graph-based model, Dynamically Fused Graph Network \citep{qiu-etal-2019-dynamically}, as our baseline model. 
The remainder of this paper is organized as follows.
\begin{itemize}[leftmargin=*]
\setlength{\itemsep}{0.05cm}
	\item In Section \ref{section: model}, we first describe our baseline model. Then, we show that the graph structure can play an important role only when the pre-trained models are used in a feature-based manner. While the pre-trained models are used in the fine-tuning approach, the graph structure may not be helpful.
	\item To explain the results, in Section \ref{sec:explain}, we point out that graph-attention \citep{velivckovic2018graph} is a special case of self-attention. The adjacency matrix based on manually defined rules and the graph structure can be regarded as prior knowledge, which could be learned by self-attention or Transformers \citep{vaswani2017attention}. 
	\item In Section \ref{sec:not necessary}, we design experiments to show when we model text as an entity graph, both graph-attention and self-attention can achieve comparable results. When we treat texts as a sequence structure, only a 2-layer Transformer could achieve similar results as DFGN.
	\item In Section \ref{sec:visual}, visualized analysis show that there are diverse entity-centered attention patterns exist in pre-trained models, indicating the redundancy of entity-based graph structure.
	\item Section \ref{discussion} gives the conclusion.
\end{itemize}

\section{The Approach}
\label{section: model}

We choose the widely used multi-hop QA dataset, HotpotQA as our testbed. We reimplement DFGN \citep{qiu-etal-2019-dynamically} and modify the usage of the pre-trained model. The model first leverage a retriever to select relevant passages from candidate set and feed them into a graph-based reader. All entities in the entity graph are recognized by an independent NER model.

\subsection{HotpotQA Dataset}
HotpotQA is a widely used large-scale multi-hop QA dataset. There are two different settings in HotpotQA. In \emph{distractor} setting, each example contains 2 gold paragraphs and 8 distractor paragraphs retrieved from Wikipedia. In \emph{full wiki} setting, a model is asked to retrieve gold paragraphs from the entire Wikipedia.
In this paper, all experiments are conducted in the \emph{distractor} setting.
\subsection{Model Description}
\noindent\textbf{Retriever.} We use RoBERTa large model \citep{liu2019roberta} to calculate the relevant score between the query and each candidate paragraphs.
We filter the paragraphs whose score is less than 0.1, and the maximum number of selected paragraphs is 3. Selected paragraphs are concatenated as context $\mathbf{C}$.

\noindent\textbf{Encoding Layer.} We concatenate the query $\mathbf{Q}$ and context $\mathbf{C}$ and feed the sequence into another RoBERTa large model.
The results are further fed into a bi-attention layer \citep{seo2016bidirectional} to obtain the representations from the encoding layer.

\noindent\textbf{Graph Fusion Block.} Given context representations $\mathbf{C}_{t-1}$ at hop $t-1$, the tokens representations are passed into a mean-max pooling layer to get nodes representations in entity graph $\mathbf{H}_{t-1}\in\mathbb{R}^{2d \times N}$, where $N$ is the number of entity. 
After that, a graph-attention layer is applied to update nodes representations in the entity graph:
{
\setlength\abovedisplayskip{10pt}
\begin{align}
	\beta^{(t)}_{i,j}&={\rm \textmd{LeakyReLU}}(\mathbf{W}_t^\top[\mathbf{h}_i^{(t-1)}, \mathbf{h}_j^{(t-1)}]) \label{eq:gat start}\\
	\alpha_{i,j}^{(t)}&=\frac{{\rm exp}(\beta_{i,j}^{(t)})}{\sum_{k\in\mathcal{N}_i}{{\rm exp}(\beta_{i,k}^{(t)})}}\\
	\mathbf{h}_i^{(t)}&={\rm \textmd{ReLU}}(\sum_{k\in\mathcal{N}_i}{\alpha_{i,k}^{(t)}\mathbf{h}_k^{(t-1)}})\label{eq: gat end}
\end{align}
\setlength\belowdisplayskip{8pt}
}
where $\mathcal{N}_i$ is the set of neighbors of node $i$. 
We follow the same Graph2Doc module as \citet{qiu-etal-2019-dynamically} to transform the nodes representations into the tokens representations. 
Besides, there are several extra modules in the graph fusion block, including query-entity attention, query update mechanism, and weak supervision.

\noindent\textbf{Prediction Layer.} We follow the same cascade structure as \citet{qiu-etal-2019-dynamically} to predict the answers and supporting sentences.

\noindent\textbf{Entity Graph Construction.} We fine-tune a pre-trained BERT base model on the dataset of the CoNLL'03 NER shared task \citep{tjong-kim-sang-de-meulder-2003-introduction} and use it to extract entities from candidate paragraphs. Connections between entities are defined as following rules:
\begin{itemize}
\setlength{\itemsep}{0.05cm}
	\item Entities with the same mention text in context are connected.
	\item Entities appear in the same sentence are connected.
\end{itemize}

\begin{table}[t]
	\small
	\centering
	\begin{tabular}{lcccccc}	
		\toprule
		\textbf{Setting} & \textbf{Joint EM} & \textbf{Joint F1} \\
		\midrule
		Baseline \citep{yang-etal-2018-hotpotqa} & 10.83 & 40.16 \\
		QFE \citep{nishida-etal-2019-answering} & 34.63 & 59.61 \\
		DFGN \citep{qiu-etal-2019-dynamically} & 33.62 & 59.82 \\
		TAP2 \citep{glass2019span} & 39.77 & 69.12 \\
		HGN \citep{fang2019hierarchical} & 43.57 & 71.03 \\
		SAE \citep{tu2019select} & \textbf{45.36} & 71.45 \\
		\midrule
		Our Model & 44.67 & \textbf{72.73} \\
		\bottomrule
	\end{tabular}
	\caption{\label{test-result}
	Results on the test set of HotpotQA.}
\end{table}

\begin{table}
	\small
	\centering
	\begin{tabular}{lcc}
		\toprule
		\textbf{Setting} & \textbf{Joint EM} & \textbf{Joint F1} \\
		\midrule
		Baseline (Fine-tuning) & 45.91 & \textbf{73.93} \\
		\quad w/o Graph & \textbf{45.98} & 73.78 \\
		Baseline (Feature-based) & 36.45 & 63.75 \\
		\quad w/o Graph & 32.26 & 59.76 \\
		\bottomrule
	\end{tabular}
	\caption{\label{ablation graph}
	Ablation of graph structure under different settings.}
\end{table}

\begin{figure*}[h]
	\centering
	\includegraphics[width = 15cm]{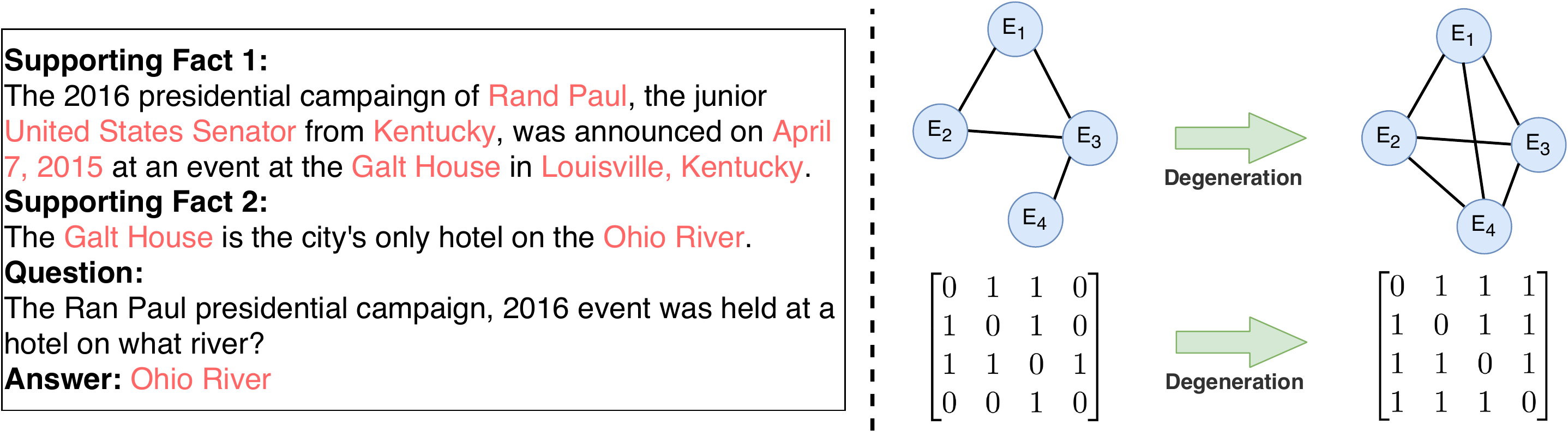}
	\caption{\label{fg: example}
	Entities in raw texts are modeled as an entity graph and handled by graph attention networks. When the entity graph are fully connected, a graph-attention layer will degenerate into a vanilla self-attention layer.
	}
\end{figure*}

\subsection{Model Results}
\label{sec:graph}
In Table~\ref{test-result}, we show the performance comparison with different models on the blind test set of HotpotQA. Our strong baseline model achieves state-of-the-art results on the official leaderboard.

In order to analyze how much the graph structure contributes to the entire model, we perform a set of ablation experiments. We remove the whole graph fusion block, and the outputs of the pre-trained model are directly fed into the prediction layer. By the reason that the main difference between our baseline model and DFGN is that we use a large pre-trained model in the fine-tuning approach instead of the feature-based approach, we perform the experiments in two different settings.  

The results are shown in Table \ref{ablation graph}. By using the fine-tuning approach, model with and without graph fusion block can reach almost equal results. When we fix parameters of the pre-trained model, the performances significantly degrade by 9\% for EM and 10\% for F1. If we further remove graph fusion block, both EM and F1 drop about 4\%.

Taken together, only when pre-trained models are used in the feature-based approach, graph neural networks can play an important role. Nevertheless, 
when pre-trained models are used as a fine-tuning approach, which is a common practice, graph structure does not contribute to the final results. In other words, the graph structure may not be necessary
 for multi-hop question answering.

\section{Understanding Graph Structure}
\label{comparing}
 Experimental results in Section \ref{sec:graph} imply that self-attention or Transformer may have superiority in multi-hop question answering. To understand this, in this section, we will first discuss the connection between graph structure, graph-attention, and self-attention. We then verify the hypothesis by experiments and visualized analysis.

\subsection{Graph Attention vs. Self Attention}
\label{sec:explain}
The key to solving the multi-hop question is to find the corresponding entity in the original text through the query. 
Then one or more reasoning paths are constructed from these start entities toward other identical or co-occurring entities. 
As shown in Figure \ref{fg: example}, previous works usually extract entities from multiple paragraphs and model these entities as an entity graph. The adjacency matrix is constructed by manually defined rules, which usually the co-occurrence relationship of entities. 
From this point of view, both the graph structure and the adjacency matrix can be regarded as task-related prior knowledge. The entity graph structure restricts the model can only do reasoning based on entities, and the adjacency matrix assists the model to ignore non-adjacent nodes in a hop. 
However, it is probably that the model without any prior knowledge can still learn the entity-to-entity attention pattern. 

In addition, considering Eq.\ref{eq:gat start}-\ref{eq: gat end}, it is easy to find that graph-attention has a similar form as self-attention. 
In forward propagation, each node in the entity graph calculates attention scores with other connected nodes. 
As shown in Figure \ref{fg: example}, graph-attention will degenerate into a vanilla self-attention layer when the nodes in the graph are fully connected. Therefore, the graph-attention can be considered as a special case of self-attention. 

\subsection{Graph Structure May Not Be Necessary}
\label{sec:not necessary}
According to the discussion above, we aimed to evaluate whether the graph structure with an adjacency matrix is superior to self-attention. 

To this end, we use the model described in Section \ref{section: model} as our baseline model. The pre-trained model in the baseline model is used in the feature-based approach. Several different modules are added between the encoding layer and the prediction layer.

\noindent\textbf{Model With Graph Structure.} We apply graph-attention or self-attention on the entity graph and compare the difference in the final results. 
In order to make a fair comparison, we choose the self-attention that has the same form as graph-attention. The main difference is that the self-attention does not keep an adjacency matrix as prior knowledge and the entities in the graph are fully connected. 
Moreover, we define that the density of a binary matrix is the percentage of `1' in it.
We sort each example in development set by the density of its adjacency matrix and divide them by different quantiles.
We evaluate how different density of the adjacency matrix affects the final results.

\noindent\textbf{Model Without Graph Structure.} In this experiment, we verify whether the whole graph structure can be replaced by Transformers. We directly feed the context representations from the encoding layer into the Transformers.

\begin{table}
	\small
	\centering
	\begin{tabular}{lcc}
		\toprule
		\textbf{Setting} & \textbf{Joint EM} & \textbf{Joint F1} \\
		\midrule
		Baseline & 32.26 & 59.76 \\
		  \quad + Graph Fusion Block & \textbf{36.45} & 63.75 \\
		  \midrule
		  \quad + Self Attention & 35.41 & 61.77 \\
		  \quad + Graph Attention & 35.79 & 61.91 \\
		 \midrule
		  \quad+ Transformer & 36.23 & \textbf{63.82} \\
		\bottomrule
	\end{tabular}
	\caption{\label{compare table}
	Performance comparison in terms of joint EM and F1 scores under different module settings.}
\end{table}

\begin{figure}
	\centering
	\includegraphics[width = 7.5cm]{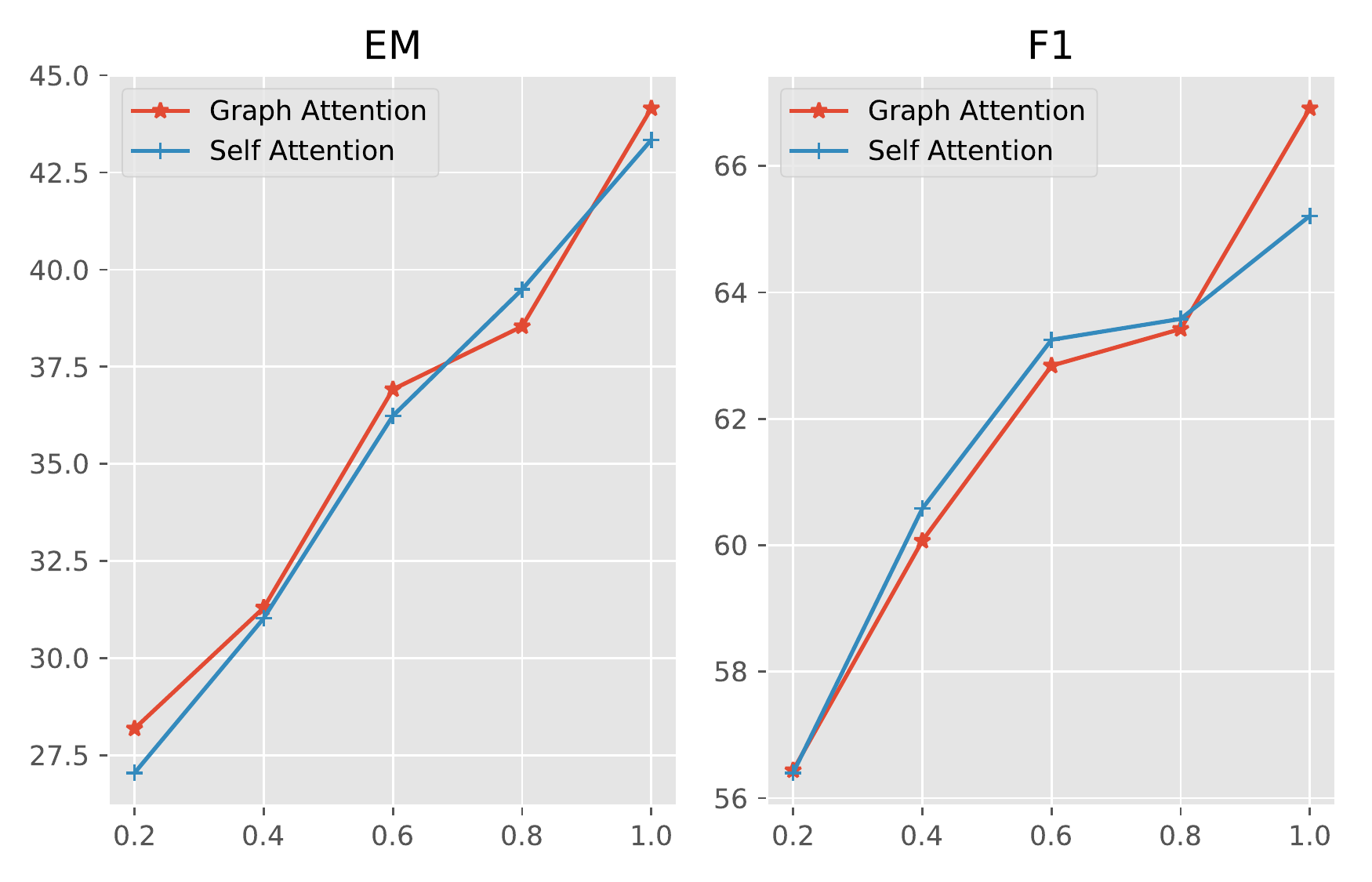}
	\caption{\label{fg: attention results}
	Results of graph-attention and self-attention on examples with different adjacency matrix density.}
\end{figure}

The experimental results are shown in Table \ref{compare table}. Compared with the baseline, the model with the graph fusion block obtains a significant advantage. 
We add the entity graph with self-attention to the baseline model, and the final results significantly improved.
Compared with self-attention, graph-attention does not show a clear advantage. 
The density of examples at different quantile are shown in Table \ref{count table}, the adjacency matrix in multi-hop QA is relatively dense, which may causes graph-attention can not make a significant difference.
The results of graph-attention and self-attention in the different intervals of density are shown in Figure \ref{fg: attention results}. 
Despite the different density of the adjacency matrix, graph-attention consistently achieves similar results as self-attention. 
This signifies that self-attention can learn to ignore irrelevant entities. 
Besides, examples with a more dense adjacency matrix are simpler for both graph-attention and self-attention, this probably because these adjacency matrices are constructed from shorter documents.
Moreover, as shown in Table \ref{compare table}, Transformers show a powerful reasoning ability. Only stacking two layers of the Transformer can achieve comparable results as DFGN.

\begin{table}
	\small
	\centering
	\begin{tabular}{p{1.1cm}p{0.5cm}<{\centering}p{0.5cm}<{\centering}p{0.5cm}<{\centering}p{0.5cm}<{\centering}p{0.5cm}<{\centering}p{0.5cm}<{\centering}}
		\toprule
		\textbf{Quantile} & 0.2 & 0.4 & 0.6 & 0.8 & 1.0 & avg \\
		\midrule
		\textbf{Density} & 18.7 & 23.6 & 29.0 & 36.8 & 100 & 28.8 \\
		\bottomrule
	\end{tabular}
	\caption{\label{count table}
	The Adjacency Matrix density at different quantiles.}
\end{table}

\subsection{Training Details}
For all experiments in this paper, the number of layers of different modules is two, and the hidden dimensions are set to 300.
In feature-based setting, all models are trained for 30 epochs with a batch size of 24.
In fine-tuning setting, models are trained for 3 epochs with a batch size of 8.
 The initial learning rate is 2e-4 and 3e-5 in the feature-based setting and fine-tuning setting respectively.

\subsection{Entity-centered Attention Pattern in Pre-trained Model}
\label{sec:visual}
Inspired by \citet{kovaleva-etal-2019-revealing}, we leverage an approximate method to find which attention head contains entity-centered attention patterns. We employ an NER model to identify tokens belong to a certain entity span. Then, for each attention head in the pre-trained model, we sum the absolute attention weights among those tokens belong to an entity and tokens not belong to an entity. The score of an attention head is the difference between the sum of weights from entities and non-entities tokens. We then average the derived scores over all the examples. Finally, the attention head with the maximum score is the desired head that contains entity-centered attention patterns.  

\begin{figure}
	\centering
	\includegraphics[width = 7.5cm]{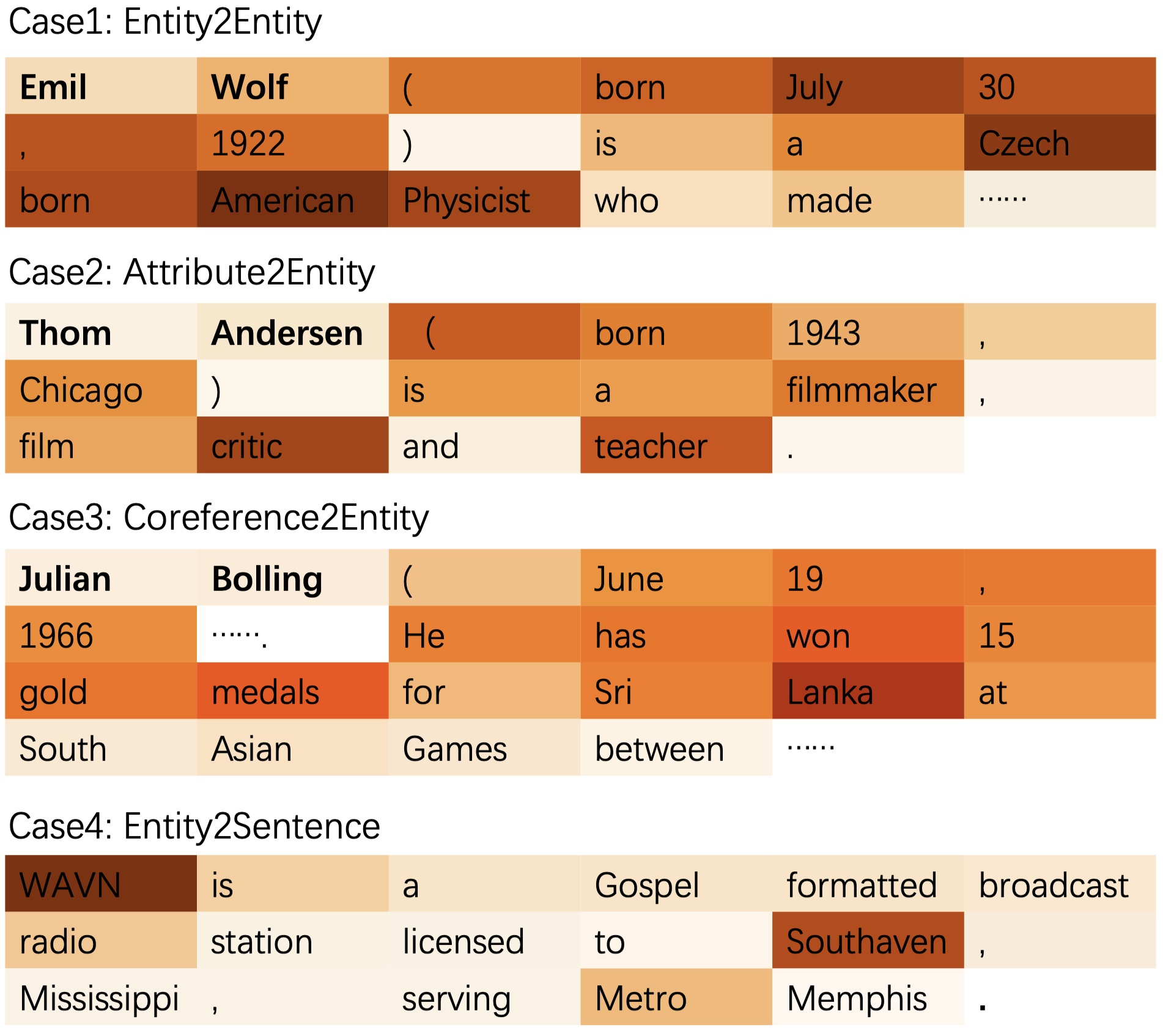}
	\caption{\label{fg: visualized analysis}
	Results of visualized attention patterns in the pre-trained model.}
\end{figure}

We find four typical attention patterns and visualized it in Figure \ref{fg: visualized analysis}. In case 1-3, we visualized the attention weights of each token attending to the subject entity. In case 4, we visualized the attention weights of each token attending to the last token of the sentence. The results show pre-trained models are pretty skillful at capturing relations between entities and other constituents in a sentence.

\noindent\textbf{Entity2Entity.} We find entity-to-entity attention pattern is very widespread in pre-trained models. In this case, `American Physicist' and 'Czech' attend to `Emil Wolf' with very high attention weights. Such attention pattern plays the same role as graph attention plays.

\noindent\textbf{Attribute2Entity.} In this case, `filmmaker', `film critic' and `teacher' obtain higher weights, indicating the occupation of `Thom Andersen'. Note that these tokens are not part of a certain entity, hence deem to be ignored by graph structure.

\noindent\textbf{Coreference2Entity.} We also find that coreference will not make the pre-trained model confused. In case 3, the entity `Sri Lanka' in second sentence attends to `Julian Bolling' in the first sentence, which means the pre-trained model understand `He' refers to `Julian Bolling' even though they belong to different sentences. 

\noindent\textbf{Entity2Sentence.} We find many entities attend to the last token of sentence. In the prediction layer, the representations of the first and last token in a sentence are combined to determine  whether a particular sentence is a supporting fact. Therefore, we suppose this is another attention pattern that entities attend to the whole sentence.

It is obvious that graph attention can not cover the last three attention patterns. Therefore, we draw a conclusion that self attention has advantages on generality and flexibility.

\section{Conclusions}
\label{discussion}
This study set out to investigate whether graph structure is necessary for multi-hop QA and what role it plays. We established that with the proper use of pre-trained models, graph structure may not be necessary. 
In addition, we point out the adjacency matrix and the graph structure can be regarded as some kind of task-related prior knowledge. 
Experiments and visualized analysis demonstrate both graph-attention and graph structure can be replaced by self-attention or Transformers. Our results suggest that future works introducing graph structure into NLP tasks should explain their necessity and superiority.

\section*{Acknowledgments}
We would like to thank all anonymous reviewers for their thorough reviewing and providing insightful comments to improve our paper.

\bibliographystyle{acl_natbib}
\bibliography{emnlp2020}

\end{document}